\documentclass[conference]{IEEEtran}
\IEEEoverridecommandlockouts

\usepackage{cite}
\usepackage{amsmath,amssymb,amsfonts}
\usepackage{algorithmic}
\usepackage{graphicx}
\usepackage{textcomp}
\usepackage{xcolor}
\usepackage{colortbl}
\usepackage{adjustbox}

\usepackage{hyperref}
\usepackage{url}
\usepackage{amsmath}
\usepackage{xcolor}
\usepackage{soul}
\usepackage{booktabs,latexsym}
\usepackage{graphicx,amsmath,epsfig,subfigure,cite,lscape,longtable}
\usepackage{times,mathptm}
\usepackage{amssymb}
\usepackage{enumerate}
\usepackage{float}
\usepackage{array}
\usepackage[figuresleft]{rotating}
\usepackage{multirow}
\usepackage{colortbl}
\newcolumntype{L}{>{\centering\arraybackslash}m{2.25cm}}
\newcolumntype{R}{>{\centering\arraybackslash}m{3.05cm}}

\def\BibTeX{{\rm B\kern-.05em{\sc i\kern-.025em b}\kern-.08em T\kern-.1667em\lower.7ex\hbox{E}\kern-.125emX}}

\usepackage{amsmath,amsfonts,bm}

\def\eqref#1{equation~\ref{#1}}

\def\1{\bm{1}}

\DeclareMathAlphabet{\mathsfit}{\encodingdefault}{\sfdefault}{m}{sl}
\SetMathAlphabet{\mathsfit}{bold}{\encodingdefault}{\sfdefault}{bx}{n}

\usepackage{multirow}
\usepackage{hyperref}
\usepackage{url}
\usepackage{soul}
\usepackage{booktabs,latexsym}
\usepackage{epsfig,subfigure,lscape,longtable}
\usepackage{times,mathptm}
\usepackage{enumerate}
\usepackage{float}
\usepackage{array}
\usepackage[figuresleft]{rotating}

\title{LSA-PINN: Linear Boundary Connectivity Loss for Solving PDEs on Complex Geometry}

\makeatletter
\newcommand{\newlineauthors}{%
  \end{@IEEEauthorhalign}\hfill\mbox{}\par
  \mbox{}\hfill\vspace{-1cm}\begin{@IEEEauthorhalign}
}
\makeatother

\author{
\IEEEauthorblockN{Jian Cheng Wong$^{*}$\thanks{*: equal contribution}}
\IEEEauthorblockA{\footnotesize\textit{Institute of High Performance Computing} \\
\textit{Agency for Science, Technology and Research}\\
Singapore \\
wongj@ihpc.a-star.edu.sg}\\
\and
\IEEEauthorblockN{Pao-Hsiung Chiu$^{*}$}
\IEEEauthorblockA{\footnotesize\textit{Institute of High Performance Computing} \\
\textit{Agency for Science, Technology and Research}\\
Singapore \\
chiuph@ihpc.a-star.edu.sg}\\
\and
\IEEEauthorblockN{Chinchun Ooi}
\IEEEauthorblockA{\footnotesize\textit{Institute of High Performance Computing} \\
\textit{Agency for Science, Technology and Research}\\
Singapore \\
ooicc@ihpc.a-star.edu.sg}\\
\newlineauthors
\IEEEauthorblockN{My Ha Dao}
\IEEEauthorblockA{\footnotesize\textit{Institute of High Performance Computing} \\
\textit{Agency for Science, Technology and Research}\\
Singapore \\
daomh@ihpc.a-star.edu.sg}\\
\and
\IEEEauthorblockN{Yew-Soon Ong}
\IEEEauthorblockA{\footnotesize\textit{School of Computer Science and Engineering} \\
\textit{Nanyang Technological University}\\
Singapore \\
asysong@ntu.edu.sg}\\
}

\begin{document}

\maketitle

\begin{abstract}
We present a novel loss formulation for efficient learning of complex dynamics from governing physics, typically described by partial differential equations (PDEs), using physics-informed neural networks (PINNs). In our experiments, existing versions of PINNs are seen to learn poorly in many problems, especially for complex geometries, as it becomes increasingly difficult to establish appropriate sampling strategy at the near boundary region. Overly dense sampling can adversely impede training convergence if the local gradient behaviors are too complex to be adequately modelled by PINNs. On the other hand, if the samples are too sparse, existing PINNs tend to overfit the near boundary region, leading to incorrect solution. To prevent such issues, we propose a new Boundary Connectivity (BCXN) loss function which provides linear \textit{local structure approximation} (LSA) to the gradient behaviors at the boundary for PINN. Our \textit{BCXN-loss} implicitly imposes local structure during training, thus facilitating fast physics-informed learning across entire problem domains with order of magnitude sparser training samples. This LSA-PINN method shows a few orders of magnitude smaller errors than existing methods in terms of the standard L2-norm metric, while using dramatically fewer training samples and iterations. Our proposed LSA-PINN does not pose any requirement on the differentiable property of the networks, and we demonstrate its benefits and ease of implementation on both multi-layer perceptron and convolutional neural network versions as commonly used in current PINN literature.
\end{abstract}

\begin{IEEEkeywords}
physics-informed neural networks, local structure approximation, boundary connectivity, partial differential equations
\end{IEEEkeywords}

\section{Introduction}

\begin{figure*}[htbp]
\begin{center}
\centerline{\includegraphics[width=0.95\linewidth]{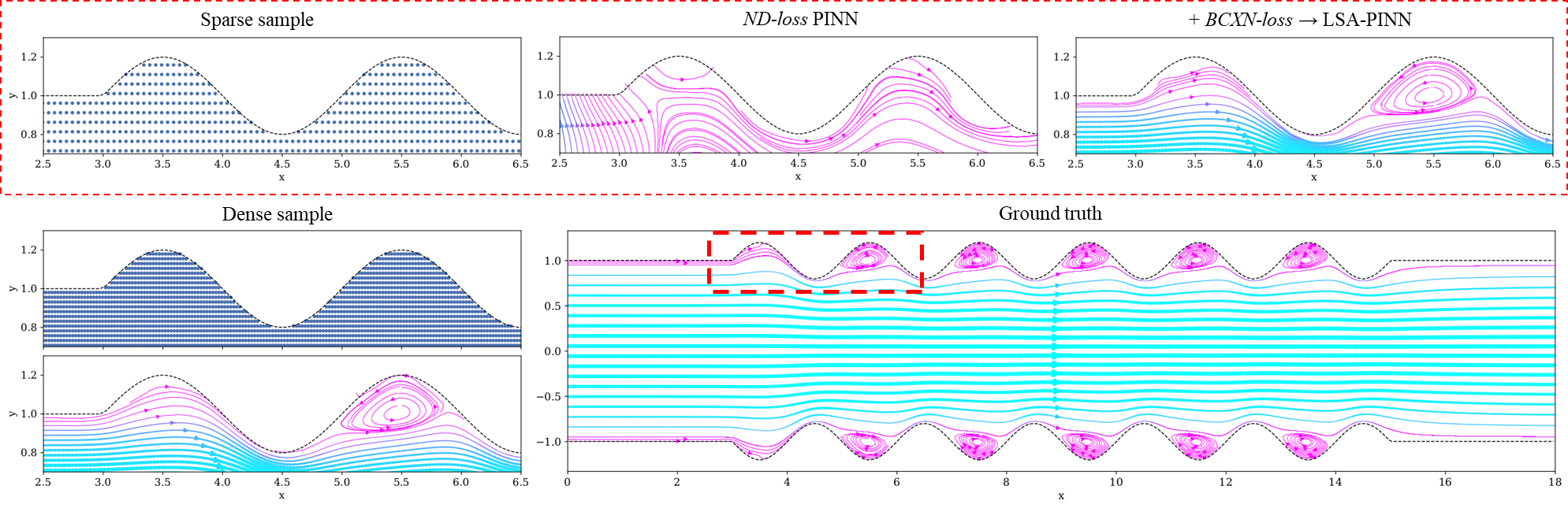}}
\caption{PINNs learning the solution of 2D N-S equations in a complex geometry (wavy channel flow problem, $Re=100$). Our LSA-PINN method can learn accurate solution with fewer (50\% less) training samples and faster speed (50\% less training iterations).}
\label{fig:fast-pinn-advantage}
\end{center}
\vspace*{-5mm}
\end{figure*}

Physics-informed neural networks (PINNs) have emerged as a promising method for learning the solution of dynamical system from the governing physics~\cite{raissi2019physics}. PINNs have recently been studied for a wide range of physical phenomena and applications across science and engineering domains — electromagnetic, fluid dynamics, heat transfer, etc~\cite{karniadakis2021physics,cuomo2022scientific}. The distinctive feature of PINNs is the use of governing physics law, typically in the form of partial differential equations (PDEs), as the learning objective. This physics-informed learning constrains the PINN from violating the underlying physics at all training points sampled from the problem domain.

Existing PINNs evaluate the PDE constraints in their training loss by either automatic differentiation (AD) or numerical differentiation (ND)-type method. \textit{AD-loss} has become a popular choice because it can exactly compute the derivative terms at any sample location directly from the input and the network weights. While both methods have their pros and cons, \textit{ND-loss} can be flexibly implemented across many different neural network (NN) architectures, including both multi-layer perceptrons (MLPs) and convolutional neural networks (CNNs), because they do not require the NN to retain differentiability, unlike AD. Recent studies ~\cite{gao2021phygeonet,fang2021high,chiu2022can} have suggested that ND-type methods and especially \textit{coupled-automatic-numerical
differentiation (CAN)-loss} ~\cite{chiu2022can} can more robustly and efficiently produce accurate solutions with fewer training samples, whereas conventional \textit{AD-loss} is prone to failure during training. This is because ND-type methods approximate high order derivatives using \emph{PINN output from neighbouring samples}, hence, they can effectively connect sparse samples into piecewise regions via these local approximations, thereby facilitating fast physics-informed learning across the entire domain with sparser samples.

When dealing with irregular geometries, it becomes difficult for existing PINNs to perfectly connect training samples in the domain’s interior to the boundary. Failing to do so can cause undesirable training failure as the PINN starts to overfit at the near boundary region. Since many PDEs of practical interest are boundary-value problems, it is desirable to have the PINNs model the correct boundary behaviors. While adding dense samples to better refine the piecewise local regions near the boundary may improve accuracy, the extent to which sampling needs to be increased is empirical, and a denser sampling strategy may adversely impede training convergence. Moreover, it may not be feasible to locally refine the near boundary sampling density for PINNs with CNN-architecture. 

Hence, we propose a loss function formulation in this work that helps provide linear approximation to the local gradient behaviour at the boundary, thereby restoring connectivity between domain boundary and near boundary interior samples. Such \textit{local structure approximation} also provides certain structural bias during PINN training to prevent inaccurate representation near the boundaries. This new boundary connectivity (\textit{BCXN})-loss function is key to a novel class of LSA-PINN method which can more efficiently learn the solution to PDEs with sparser training samples, regardless of domain geometry; see example in Fig.~\ref{fig:fast-pinn-advantage}. In addition, this method can be jointly implemented with other PINN advances in optimization such as loss balancing, domain decomposition and adaptive sampling. 

In the rest of this work, we present comprehensive experiments to demonstrate i) the flexible implementation of LSA-PINN method for both MLP and CNN architectures; and ii) the effectiveness of LSA-PINN for learning multiple complex fluid dynamical systems, including for meta-modelling. Compared to conventional PINNs with the \textit{AD-} and \textit{ND-loss}, our LSA-PINNs with linear \textit{BCXN-loss} are shown to be capable of tackling challenging PDE problems while using sparser training samples, hence expanding the exciting potential of PINNs for learning complex, real-world dynamics.

\section{Related Work}

\textbf{Efficient sampling in PINNs.} The theoretical limit of physics-informed loss learning in relation to training samples has been provided by prior studies ~\cite{lu2021machine,mishra2022estimates}. To improve PINN training speed for practical applications ~\cite{markidis2021old}, several studies have focused on efficient sampling strategies such as importance sampling, adaptive sampling, and sequential sampling to reduce the amount of training samples required during PINN training ~\cite{anitescu2019artificial,mcclenny2020self,wight2020solving,nabian2021efficient,lu2021deepxde,lye2021iterative,daw2022rethinking,mattey2022novel, wu2023comprehensive}. Domain decomposition and parallelization strategies have also been explored to speed up training ~\cite{jagtap2020conservative,jagtap2021extended,shukla2021parallel,dong2021local,li2019d3m}. Our LSA-PINN differs from these works in that we make physics-informed learning more robust in the sparse sample regime via a newly-proposed \textit{BCXN-loss}.

\textbf{CNN architecture and numerical differentiation (ND)-type loss for PINNs.} CNN-based formulation allows us to design and train larger, more powerful networks, hence it has potential to be scalable for more complex, large-scale PDE problem~\cite{wandel2020learning,gao2021phygeonet,wandel2021teaching,ranade2021discretizationnet,wandel2022spline,ren2022phycrnet}. However, cumbersome coordinate transformations may need to be performed to better handle irregular domains. CNN-architecture PINNs usually utilize ND methods for computing the PDE loss on their input grid, which is much cheaper to compute as compared to AD-based loss ~\cite{gao2021phygeonet}. Hybrid frameworks that couple both AD- and ND-type loss have also been proposed for MLP architecture PINNs to unify the advantages of both methods ~\cite{jagtap2020conservative,mitusch2021hybrid,fang2021high,chiu2022can}. Our method further augments the efficiency and applicability of ND-type loss on irregular geometries.

\textbf{Enforcing BC constraint in PINN loss.} It is important for the PINNs to prioritize learning the correct boundary behaviors~\cite{shin2020convergence, wang2022respecting}. The BCs are usually enforced into the PINN loss as a soft constraint via penalty method. Various strategies have been explored in this context to dynamically calibrate the relative important between PDE and BC constraints during PINN training~\cite{elhamod2020cophy,bischof2021multi,jin2021nsfnets,thanasutives2021adversarial,maddu2022inverse,de2022mopinns,van2022optimally,xiang2022self,huanguniversal, wang2021understanding}. There are also other approaches to bypass the loss balancing issue. For example, one can devise an ansatz function such that the BCs are exactly satisfied by construction or implicitly formulate the BC constraints into the PDE loss~\cite{lagaris1998artificial,lagaris2000neural,mcfall2009artificial,berg2018unified,nabian2019deep,karumuri2020simulator,wang2020mesh}, leaving only single loss term from PDE residuals to be optimized. Another example is the use of augmented Lagrangian method to impose the BC constraints into PINN loss~\cite{lu2021physics}. Our method implicitly incorporates BCs into the PDE loss via a different strategy.

\section{Preliminary}

\subsection{Governing physics - incompressible Navier-Stokes (N-S) equations}

Fundamentally, PINNs have been shown to be applicable to many different physical systems. The present study focuses on learning fluid dynamics with PINNs. We consider fluid problems where the governing physics are the steady-state, incompressible N-S equations derived from the conservation of mass and momentum:
\begin{subequations} 
\label{eq:ns_eqn}
\begin{align}
&\nabla\cdot\vec{u} = 0 \\
&\left(\vec{u}\cdot\nabla\right)\vec{u} = Re^{-1}\Delta\vec{u} - \nabla p
\end{align}
\end{subequations} 
In the above PDEs, the primitive variables $\vec{u}$ and $p$ are velocity vector and pressure while Reynolds number ($Re$) represents the ratio of inertial to viscous forces. $\vec{u}$ consists of 2 components $\left(u,v\right)$ for a 2D case. While this work uses Cartesian coordinates, this formulation is extendable to other coordinate systems.

\subsection{MLP- and CNN-architecture PINNs}

We use a fully connected DNN architecture (e.g., MLP, CNN) to represent the solution of the dynamical process \textbf{\textit{U}} $= [\vec{u},p]^{T}$. For MLPs, the input $\vec{x}$ is a point coordinate in spatial domain. For CNNs, the input \textbf{\textit{X}} is a tensor with a fixed shape, comprising the entire (discretized) spatial domain. The accuracy of the PINN outputs \textbf{\textit{U}}($\vec{x}$;\textbf{\textit{w}}) given input $\vec{x}$ is determined by the network parameters \textbf{\textit{w}}, which are optimized w.r.t. the PINN loss function during training. The PINN loss function is defined as the composition of a PDE loss component $\left(L_{PDE}\right)$ and a BC loss component $\left(L_{BC}\right)$:
\begin{subequations} 
\label{eq:loss_fn}
\begin{align}
L_{PINN} &= \lambda_{PDE} L_{PDE} + \lambda_{BC} L_{BC}  \\
\begin{split}
L_{PDE} &= \big\|\nabla\cdot\vec{u}(\vec{x};\textbf{\textit{w}})\big\|_{\Omega}^2 + \big\|\left(\vec{u}(\vec{x};\textbf{\textit{w}})\cdot\nabla\right)\vec{u}(\vec{x};\textbf{\textit{w}}) \\
& - Re^{-1}\Delta\vec{u}(\vec{x};\textbf{\textit{w}}) + \nabla p(\vec{x};\textbf{\textit{w}}) \big\|_{\Omega}^2 \end{split} \\
L_{BC} &= \big\|B[u(\vec{x};\textbf{\textit{w}})] - \vec{u}(\vec{x}) \big\|_{\partial\Omega}^2 
\end{align}
\end{subequations} 
The PDE loss penalizes deviation from governing N-S equations for the PINN output \textbf{\textit{U}}($\vec{x}$;\textbf{\textit{w}}) over the fluid domain $\vec{x}\in\Omega$, whereas the BC loss penalizes deviation from the desired Dirichlet boundary condition $\vec{u}_{BC}(\vec{x})$ which constrains the velocity field at the domain boundary $\vec{x}\in\partial\Omega$. The relative weights $\lambda_{PDE}$ and $\lambda_{BC}$ in Eq.(\ref{eq:loss_fn})a control the trade-off between different loss components.

The PDE and BC loss components are defined over a continuous domain, but for practical reasons, we evaluate the discretized PINN loss from a finite set of $n$ samples $D=\{( \vec{x}_{i} )\}_{i=1}^{n}$ during training. The CNN-architecture PINNs naturally acquire an input with equidistantly spaced grid where all the samples are coming from. For the MLP-architecture PINNs, it is also convenient to design training samples based on equidistantly spaced grid, then $m (< n)$ samples can be randomly drawn to compute the PINN loss during each stochastic gradient descent (SGD) mini-batch training iteration.

Although it is ideal for a PINN to satisfy the PDE constraint on a very dense set of samples, i.e., $\Delta\vec{x} \rightarrow0$ where $\Delta\vec{x}$ is the distance between two adjacent sample points, training such PINN may not be computationally feasible even with state-of-the-art SGD variants. For a more complex physics problem, training on dense samples is correspondingly more computationally expensive, and is also more likely to become unsuccessful due to adverse impact by some local region with very complex gradient behaviours, e.g., abnormally high gradient~\cite{fuks2020limitations,michoski2020solving,ramabathiran2021spinn,ji2021stiff,lucor2021physics,huanguniversal}. Given the same problem, training PINNs on sparser samples means fewer restrictions (the PDE constraint that needs to be satisfied) and an increased likelihood of avoiding problematic gradient points, thereby accelerating training. The \emph{relaxation of PDE constraint via reducing the sample density makes learning physically meaningful solution feasible}, although it may introduce some approximation error to the solution. Note that there is still an open question as to how best balance sample density and solution accuracy. Methods that can consistently produce more accurate solutions with sparse training samples like LSA-PINN are hugely advantageous for learning complex, real-world dynamics.

\subsection{PINN learning challenge on sparse samples}

When samples are sparse relative to the local complexity, a highly flexible, over-parameterized PINN can become “overfit”, leading to incorrect solution even as the PDE constraint is fulfilled at all sample points (i.e., the convergence is good). For \textit{AD-loss} PINN, this phenomenon can occur at any location within the computation domain. On the contrary, \textit{ND-loss} are more robust to sample density, because ND-type methods approximate the derivative terms by using \emph{output from neighbouring samples}, e.g., a finite difference-type stencil, which connects sparse samples into piecewise continuous regions to facilitate fast training across the entire problem domain. The use of lower-order \textit{local structure approximation} for the derivative terms also provides certain structural bias for PINNs to reduce overfitting. For example, abnormally high gradients are regularized so that it is easier for PINNs to learn the solution, with the accuracy being dependent on the approximation employed.

\begin{figure}[!htbp]
\begin{center}
\mbox{\subfigure[\scriptsize Stencil points connect to boundary]{\includegraphics[height=2cm,keepaspectratio]{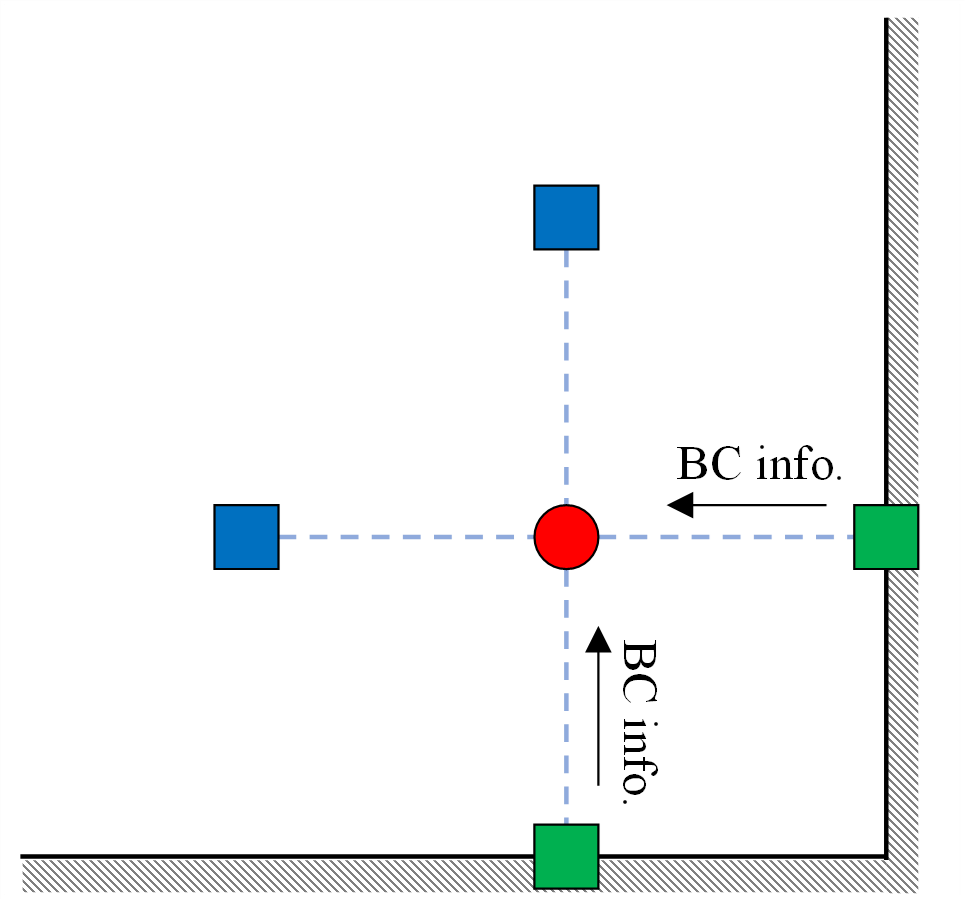}}
\hspace*{5mm}
      \subfigure[\scriptsize Stencil points fall outside domain]{\includegraphics[height=2cm,keepaspectratio]{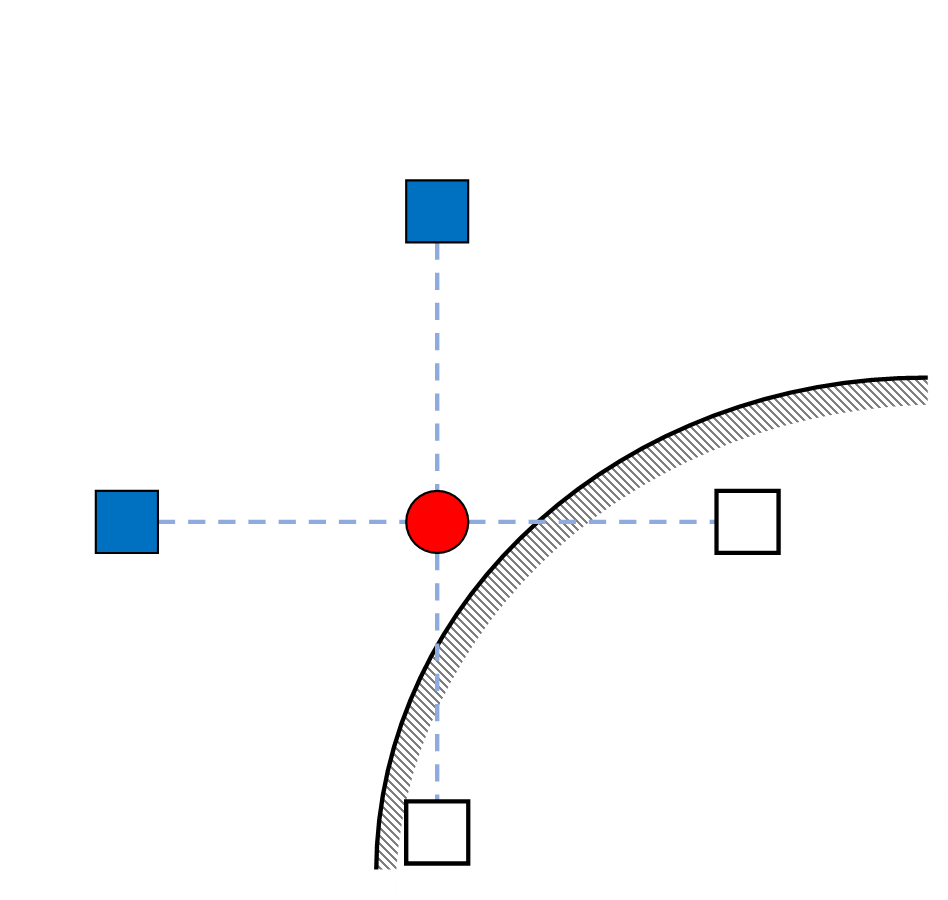}}
\hspace*{5mm}
      \subfigure[\scriptsize Extrapolation along normal direction]{\includegraphics[height=2cm,keepaspectratio]{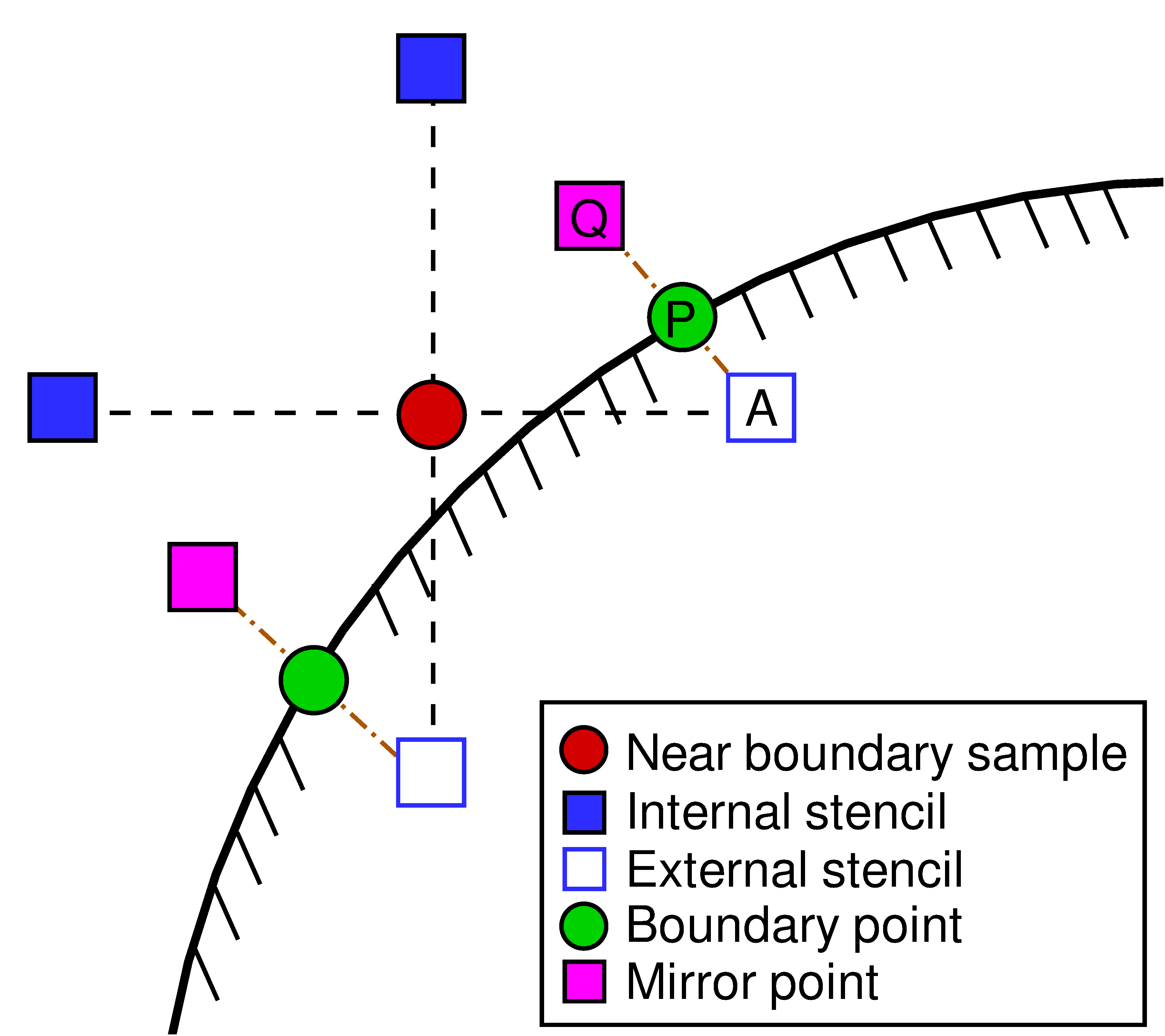}}}
\caption{(a)-(b) Schematic of a near-boundary training sample (red circle) and its stencil points (square) typically used for evaluating PDE constraint in \textit{ND-loss}. BC information is passed to stencils in fluid domain (blue square) (a) successfully in regular domain; and (b) unsuccessfully in an irregular domain due to under-defined stencils (unfilled square) outside the domain. (c) Schematic of related definitions under the present LSA-PINN framework. \textit{BCXN-loss} enforces a constraint across 3 points: external stencil point A, boundary point P, and mirror point Q inside fluid domain.}
\label{fig:stencil-schematic}
\end{center}
\vspace*{-5mm}
\end{figure}

However, \textit{ND-loss} can still exhibit training failure if the near-boundary training samples do not perfectly connect to the domain boundary through their stencils (as illustrated in Fig.~\ref{fig:stencil-schematic}). Their outputs essentially act as free parameters during optimization for satisfying the PDE constraint at near-boundary samples, and the PINN training tends to “overfit” the near-boundary region to more easily jointly satisfy the PDE and BC constraints, leading to incorrect solution. \emph{This becomes a major issue for ND-loss when dealing with irregular geometries}, because it is almost impossible to design sample locations such that their stencils perfectly connect the domain boundary to other training samples.

One potential solution is to increase the sampling density near the boundary, i.e., $\Delta\vec{x} \rightarrow0$, to avoid such over-fitting. This however increases the complexity of the optimization problem and can cause adverse effects on training. It is also very challenging to locally refine the sampling density for PINNs with CNN-architecture. To resolve this issue without increasing the sample density, we propose \textit{BCXN-loss} to impose \textit{local structure approximation} to govern the near boundary gradient behaviour and restore connectivity between domain boundary and near-boundary samples. 

\section{Method}
\label{methods}
\subsection{Enforcing linear constraint at near-boundary samples}
This section outlines the proposed LSA-PINN method. Here we denote the external (out-of-domain) stencil of a near-boundary sample as ES point(s). As per Fig.~\ref{fig:stencil-schematic}c, the boundary condition $\vec{u}_{BC}$ and field value $\vec{u}_{MI}$ are defined at the boundary point P and a chosen mirror point Q inside the domain, while the field value $\vec{u}_{ES}$ at ES point A is the value to be determined. 

By employing Taylor series expansion on points A and Q relative to point P along $\overline{AQ}$ with respect to local coordinate $n$ (normal direction of surface), the following can be derived:
\begin{subequations}
\label{eq:bc_relation}
\begin{align}
&\vec{u}_{ES} = \vec{u}_{BC} + \overline{AP}\frac{\partial\vec{u}_{BC}}{\partial n} + \frac{\overline{AP}^2}{2}\frac{\partial^{2}\vec{u}_{BC}}{\partial n^{2}} + O(\overline{AP}^3) \\
&\vec{u}_{MI} = \vec{u}_{BC} - \overline{PQ}\frac{\partial\vec{u}_{BC}}{\partial n} + \frac{\overline{PQ}^2}{2}\frac{\partial^{2}\vec{u}_{BC}}{\partial n^{2}} + O(\overline{PQ}^3) 
\end{align}
\end{subequations} 
where $\overline{AP}$ and $\overline{PQ}$ are the distances between A and P and between P and Q, and $\overline{AQ} = \overline{AP} + \overline{PQ}$. The field value $\vec{u}_{ES}$ at point A can then be derived as: 
\begin{equation} 
\label{eq:bc_eqn}
\vec{u}_{ES} = \vec{u}_{MI} + (\vec{u}_{BC} - \vec{u}_{MI})\times\frac{\overline{AQ}}{\overline{PQ}} + O(\overline{AQ}^2)
\end{equation} 
From the above equation, it can be seen that $\vec{u}_{ES}$ is linearly constrained by $\vec{u}_{BC}$ and $\vec{u}_{MI}$. While the mirror point Q can be conveniently chosen at the center of stencil, i.e., the near-boundary sample location, we choose mirror point Q along the normal direction of boundary P (ref. to Fig.~\ref{fig:stencil-schematic}c) as it is more general and performs better in our experiments. Hence, $\overline{AP} = \overline{PQ}$, and Eq. (\ref{eq:bc_eqn}) becomes:
\begin{equation} 
\label{eq:bc_normal}
\vec{u}_{ES} = 2\vec{u}_{BC} - \vec{u}_{MI}
\end{equation}
A similar process can be applied for the derivation of corresponding linear constraints for Neumann- and Robin-type boundary conditions. 

\subsection{Boundary connectivity (\textit{BCXN})-loss with direct forcing} \label{df-bcxn}

We can \emph{directly apply Eq. (\ref{eq:bc_normal}) to compute the field value $\vec{u}_{ES}$ for any external stencil point} during the evaluation of PDE constraint on near-boundary samples with ND-type schemes. This direct forcing approach has an association to the direct forcing immersed boundary methods in numerical computing. We can then use the following \textit{BCXN-loss}:
\begin{equation} 
\label{eq:bcxn_loss}
L_{LSA-PINN} = L_{PDE(BCXN)}
\end{equation}
as our LSA-PINN's PDE loss term whereby the PDE constraints on near-boundary samples are modulated by the local linear constraint. In such implementation, the BCs are implicitly infused into the training loss, obviating the BC loss term.

\subsection{Procedure to evaluate the field value $\vec{u}_{MI}$ at mirror point} \label{mirror-point}
Note that PINN methods in general require a routine to sample from inside the geometry of interest as well as the boundary. As compared to conventional PINNs, LSA-PINN involves additional steps. We first describe the steps to compute the mirror point location (along normal direction):
\begin{itemize}
\item \textbf{Determine whether a stencil is external.} In this study, we first construct a level set function $\phi$ of the geometry of interest ~\cite{osher2003geometric}, where the shortest distance $\overline{AP}$ between the stencil point A and boundary point P can be found. From the sign of level set function, we can determine whether a stencil is external.
\item \textbf{Compute mirror point location for external stencil.} Once $\overline{AP}$ is determined, we compute the location of mirror point Q inside the fluid domain, with the distance $\overline{AP} = \overline{PQ}$.
\end{itemize}
In practice, given a fixed set of training samples, all the external stencil points and their mirror locations can be pre-computed. On the other hand, the field value $\vec{u}_{MI}$ at mirror points depends on the LSA-PINN’s output and are evaluated inside a training iteration:
\begin{itemize}
\item \textbf{MLP-architecture.} The $\vec{u}_{MI}$ value can be directly obtained by evaluating $\vec{u}_{MI}(\vec{x};\textbf{\textit{w}})$ at the mirror point.
\item \textbf{CNN-architecture.} When the location of mirror point does not coincide with the CNN grid, the following inverse-distance-weighted interpolation function can be utilized to obtain the $\vec{u}_{MI}$ value at mirror point ~\cite{chiu2021development}.
\end{itemize}
\begin{subequations} 
\label{eq:interpolations}
\begin{align}
&u(\Tilde{x},\Tilde{y}) = \Sigma u(x,y) \phi '\Bigl(\frac{x-\Tilde{x}}{h}\Bigr) \phi '\Bigl(\frac{y-\Tilde{y}}{h}\Bigr) \\
&\phi '\Bigl(\frac{x-\Tilde{x}}{h}\Bigr) = \phi\Bigl(\frac{x-\Tilde{x}}{h}\Bigr) / \Sigma\phi \Bigl(\frac{x-\Tilde{x}}{h}\Bigr) \\
&\phi '\Bigl(\frac{y-\Tilde{y}}{h}\Bigr) = \phi\Bigl(\frac{y-\Tilde{y}}{h}\Bigr) / \Sigma\phi \Bigl(\frac{y-\Tilde{y}}{h}\Bigr) \\
&\phi(r) = 
\begin{cases}
    \frac{1}{8}\Bigl(3 - 2|r| + \sqrt{1 + 4|r| + 4r^{2}}\Bigr), & |r| \leq 1 \\
    \frac{1}{8}\Bigl(5 - 2|r| - \sqrt{-7 + 12|r| - 4r^{2}}\Bigr), & 1 < |r| \leq 2 \\
    0,              & |r| > 2
\end{cases}
\end{align}
\end{subequations} 
$(\Tilde{x},\Tilde{y})$ is the location of mirror point to be interpolated, while $(x,y)$ is the location of training samples inside the fluid domain. In this study, $h$ is chosen as $\Delta x$, which leads to a $2\Delta x$ interpolation radius. Once $\vec{u}_{MI}$ and $\vec{u}_{BC}$ are evaluated, we can compute $\vec{u}_{ES}$ based on Eq. (\ref{eq:bc_normal}) during training.

\section{Results}

\begin{figure*}[htbp]
\begin{center}
\centerline{\includegraphics[width=0.95\linewidth]{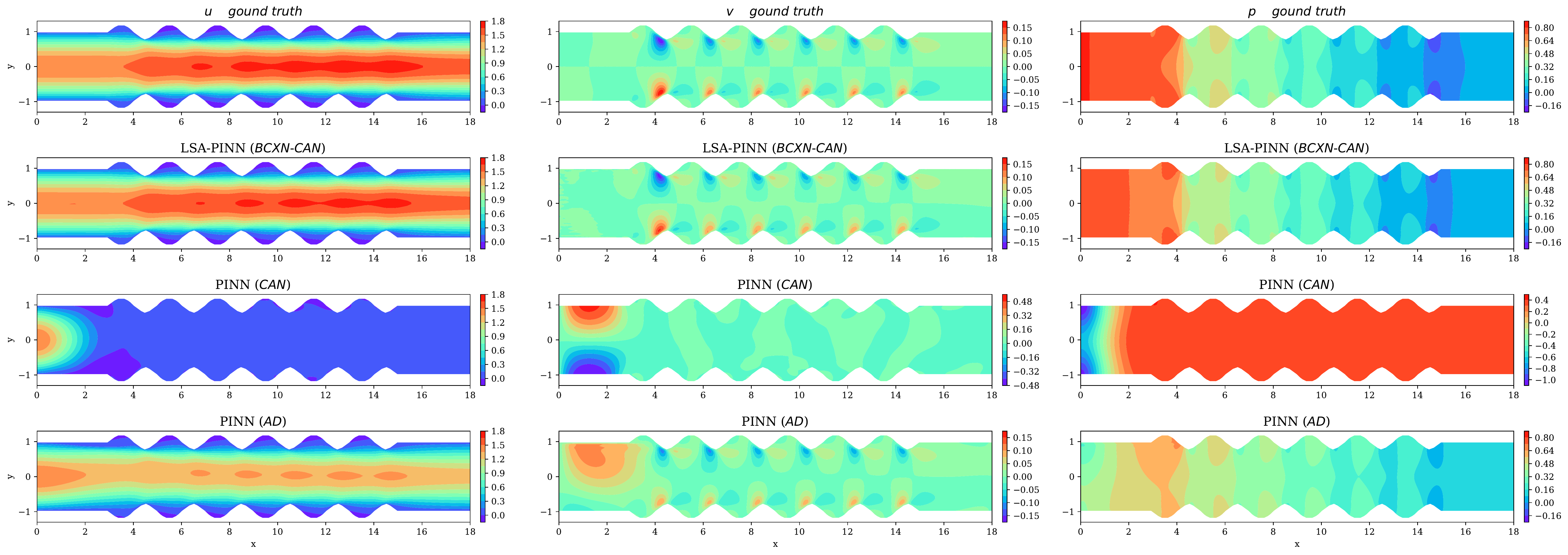}}
\caption{Comparison of $u$-, $v$-velocity and pressure $p$ contour between the ground truth and PINN solutions (median MSE from 10 runs), for the wavy channel flow problem, $Re=100$.}
\label{fig:wavy-channel-show}
\end{center}
\vspace*{-5mm}
\end{figure*}

In this work, we study the performance of the proposed LSA-PINN for learning multiple complex fluid dynamical systems, using both MLP and CNN architectures. We further demonstrate that LSA-PINN is broadly applicable to different ND-type training losses, including ND-type (\textit{BCXN-ND}) and CAN-type (\textit{BCXN-CAN}) schemes. 

The LSA-PINN (\textit{BCXN-ND-}, \textit{BCXN-CAN-loss}) is compared with baseline PINN (\textit{ND-}, \textit{CAN-}, \textit{AD-loss}) models, based on mean squared error (MSE) of the respective PINN solutions' velocity vector $\vec{u}$ relative to ground truth. To facilitate comparison, identical network architectures, initialization distribution, and training settings are employed. Relevant settings are summarized in Table~\ref{tab01} and Table~\ref{tab02} in Appendix \ref{table-settings}. For each experiment, we perform and present results based on 10 independent runs with different initialization. The ground truth solutions are obtained by an in-house numerical solver based on the improved divergence-free condition compensated (IDFC) method~\cite{chiu2018improved}. The 3 fluid dynamics test cases are:
\begin{itemize}
    \item \textbf{2D semi-circle lid-driven cavity flow, $Re=1000$.} The fluid flow inside the semi-circle cavity is driven by a lid velocity $u_{lid}=1,v_{lid}=0$ at the top wall. No-slip condition $(u=v=0)$ is applied to the other wall. There is a primary eddy near the cavity's center. 
    \item \textbf{2D lid-driven cavity flow, $Re=1000$.} The fluid flow inside a $1 \times 1$ unit square cavity is driven by a lid velocity $u_{lid}=1,v_{lid}=0$ at the top wall. No-slip condition is applied to the other walls. The lid-driven cavity flow has been widely chosen as a benchmark case for many numerical methods due to the complex physics encapsulated within. At $Re=1000$, there is a primary eddy near the center of the cavity, with secondary and tertiary eddies at the bottom-right and bottom-left regions. 
    \item \textbf{2D wavy channel flow, $Re=100$.} The fluid flow passing through a long wavy channel is studied. The inlet profile at left boundary is defined as $u(0,y)=-\frac{3}{2} y^{2}+\frac{3}{2},v=0$. A non-slip condition is applied to the top and bottom walls, while outlet boundary conditions $(\frac{\partial u}{\partial x}=\frac{\partial v}{\partial x}=p=0)$ are applied to the right boundary.
\end{itemize}

\subsection{MLP-architecture LSA-PINN \label{mlp-pinn}} 

In the MLP-architecture experiment, we evaluate the LSA-PINN's performance and compare them to other \textit{AD-}, \textit{CAN-}, and \textit{ND-loss} baseline methods previously published for the MLP architecture. All models were trained on a relatively sparse set of training sample (4,230, 10,400, and 15,200 points for semi-circle lid-driven cavity, lid-driven cavity, and wavy channel flow problems, respectively) for 300,000 iterations using Adam algorithm. The average MSE and relative L2 error across multiple independent runs are summarized in Table \ref{tab: mlp-error-all}. 

\begin{table}[htbp]
\caption{Model Performance Comparisons for MLP-architecture PINNs}
\begin{center}
\begin{adjustbox}{width=.48\textwidth}
\begin{tabular}{cccc}
\hline
\textbf{Problem} & \textbf{Model} & \textbf{MSE} & \textbf{Rel. Error} \\
\hline
\multirow{6}{*}{\parbox{3cm}{\centering Semi-circle lid-driven cavity flow, $Re = 1000$}} & {\textit{BCXN-CAN}} & {$6.21 \times 10^{-5}$} & {$3.55 \times 10^{-2}$} \\
\cline{2-4}
{} & {\textit{BCXN-ND}} & {$1.61 \times 10^{-4}$} & {$5.95 \times 10^{-2}$} \\
\cline{2-4}
{} & {\textit{CAN}} & {$2.82 \times 10^{-2}$} & {$7.62 \times 10^{-1}$} \\
\cline{2-4}
{} & {\textit{ND}} & {$3.91 \times 10^{-2}$} & {$8.56 \times 10^{-1}$} \\
\cline{2-4}
{} & {\textit{AD (SIREN)}} & {$1.74 \times 10^{-2}$} & {$5.95 \times 10^{-1}$} \\
\cline{2-4}
{} & {\textit{AD} (Vanilla)} & {$5.46 \times 10^{-2}$} & {$9.83 \times 10^{-1}$} \\
\hline
\multirow{3}{*}{\parbox{3cm}{\centering Lid-driven cavity flow, $Re = 1000$}} & {\textit{BCXN-CAN}} & {$5.10 \times 10^{-5}$} & {$3.38 \times 10^{-2}$} \\
\cline{2-4}
{} & {\textit{CAN}} & {$4.23 \times 10^{-2}$} & {$9.77 \times 10^{-1}$} \\
\cline{2-4}
{} & {\textit{AD (SIREN)}} & {$3.47 \times 10^{-2}$} & {$8.87 \times 10^{-1}$} \\
\hline
\multirow{3}{*}{\parbox{3cm}{\centering Wavy channel flow, $Re = 100$}} & {\textit{BCXN-CAN}} & {$1.34 \times 10^{-4}$} & {$7.28 \times 10^{-2}$} \\
\cline{2-4}
{} & {\textit{CAN}} & {$5.96 \times 10^{-1}$} & {$2.55 \times 10^{0}$} \\
\cline{2-4}
{} & {\textit{AD (SIREN)}} & {$3.52 \times 10^{-2}$} & {$7.19 \times 10^{-1}$} \\
\hline
\end{tabular}
\end{adjustbox}
\begin{enumerate}[(I)]

\item 
Implementation for \textit{CAN-}PINN models based on previous work~\cite{chiu2022can}.
\item 
Implementation for \textit{SIREN-}PINN models based on previous work~\cite{sitzmann2020implicit,wang2021eigenvector,wong2022learning}.
\item
Implementation for vanilla PINN models based on previous work~\cite{raissi2019physics}. 
\end{enumerate}

\label{tab: mlp-error-all}
\end{center}
\end{table}

The results indicate that our present LSA-PINN method can successfully learn a good solution with low MSE across a diverse set of problems with irregular domains and complex physics. All 3 test cases show noticeable improvement in solution accuracy ($>$2 orders of magnitude lower $\vec{u}$ MSE) with the LSA-PINN method. The baseline PINNs fail to produce a reasonable solution, i.e., they cannot properly learn the correct flow with current sampling density and training iteration, see example in Fig.~\ref{fig:wavy-channel-show}.

\begin{figure}[H]
\begin{center}
\centerline{\includegraphics[width=0.5\textwidth,keepaspectratio]{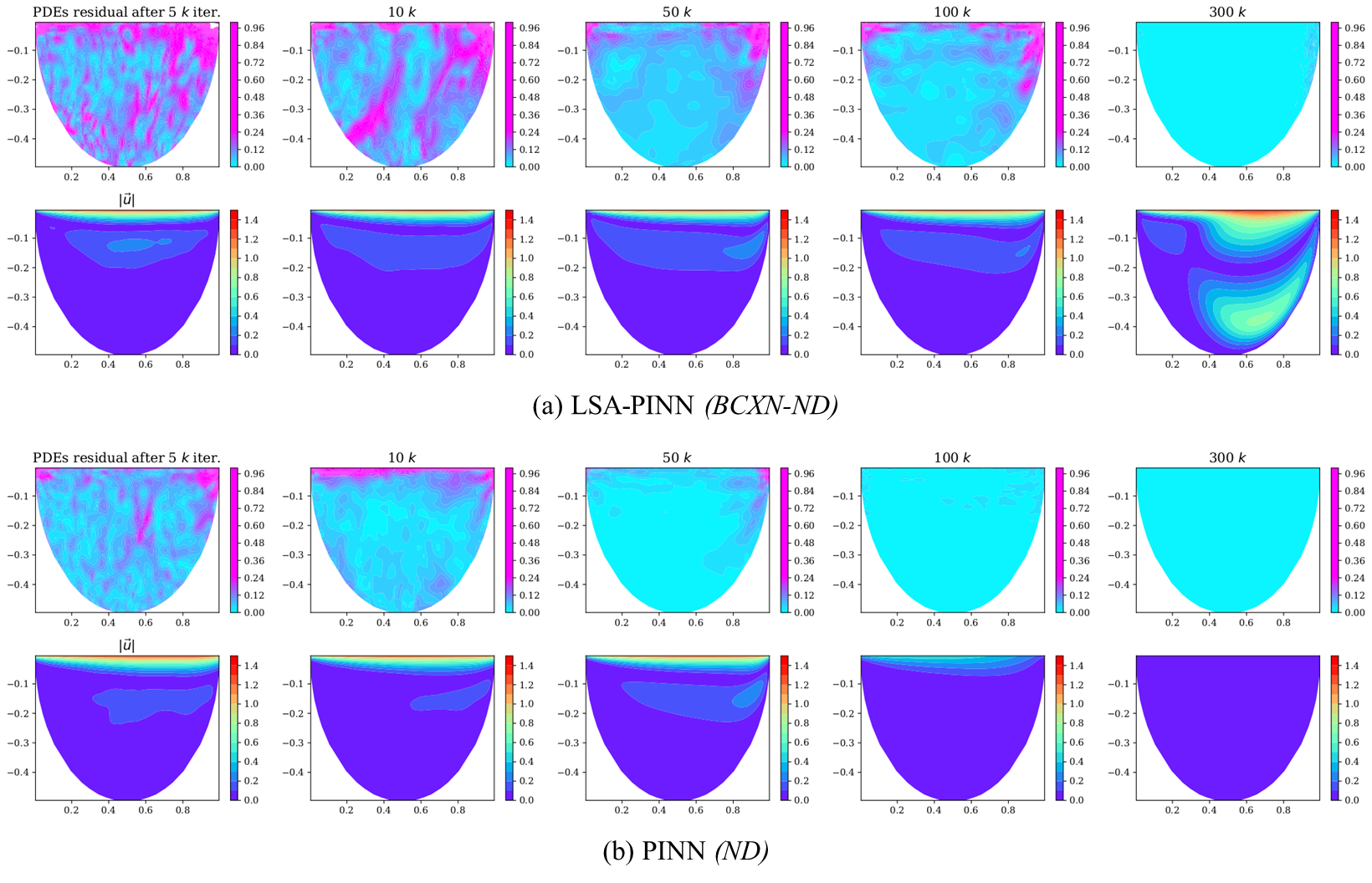}}
\vspace*{-3mm}
\caption{Contour plots of PDE error residual and velocity for (a) LSA-PINN (\textit{BCXN-ND}) and (b) PINN (\textit{ND}), for the semi-circle lid-driven cavity flow problem, $Re=1000$.}
\label{fig:BCXN-error-prop}
\end{center}
\vspace*{-6mm}
\end{figure}

\textbf{Case study of information propagation.} Information propagation from the boundary to the inner domain has been suggested to be critical for successful PINN modelling ~\cite{wang2022respecting, daw2022rethinking}. Hence, we investigate the evolution of the PDE residual across multiple training runs for the semi-circle lid-driven cavity with and without \textit{BCXN-loss} to better understand the impact of enhanced connectivity and \textit{local structure approximation} imposed by the \textit{BCXN-loss}. Illustrative contour plots of the PDE residuals and velocity at various training iterations are provided in Fig. \ref{fig:BCXN-error-prop}.

As described in the previous sections, the \textit{ND} methods typically fail to converge to a good model even after 300,000 training iterations but this can be improved via the addition of \textit{BCXN-loss} during training. The velocity contours clearly illustrate the improved convergence. In addition, there is a striking difference in the distribution of PDE residuals across the domain during the early stages of training. In a typical \textit{ND}-PINN, the PDE residuals rapidly reduce to zero, especially in the parts of the domain far away from the boundary condition. This also corresponds to the \textit{ND-loss} PINN velocity profile converging to a wrong solution, and is indicative of the information from the boundary condition being lost within the domain. On the other hand, the PDE residuals within the domain persist for a much longer period when the \textit{BCXN-loss} is included, which is indicative of information from the top boundary being persistently transmitted to the inner domain until the LSA-PINN model converges. This is also consistent with the physical dynamics of the system, as we note that the lid-driven cavity flow, as the name suggests, is heavily dependent on the top boundary's motion to drive the dynamical flow within the entire domain. 

\textbf{Case study of samples with different boundary connectivity.} The lid-driven cavity test case has a regular domain, but we chose Cell-centered (whereby near-boundary points need support points outside the domain) sample locations to show that the proposed method can be useful for this kind of scenarios. To further illustrate the impact of having near-boundary stencil points outside of the domain, we also investigate the impact of Cell-centered sampling points against Node-centered (whereby near-boundary points use support points that are coincident with the domain boundary). These two sets of sampling points are tested on a lid-driven cavity flow at $Re = 1000$ with different batch sizes and point sampling density for the \textit{BCXN-CAN}, \textit{CAN} and \textit{AD} models, and are trained for $300,000$ iterations each. The MSE and L2 relative errors are collated across multiple independent runs and tabulated in Table \ref{tab: ldc-sampling-effect}. A qualitatively good solution for this problem corresponds to an MSE of $1 \times 10^{-3}$ and below.

\begin{table}[htbp]
\caption{Model Performance for Different Sampling Densities. [C] and [N] refer to cell- and node-centered sample arrangements.}
\begin{center}
\begin{adjustbox}{width=.49\textwidth}
\begin{tabular}{|c|c|c|c|c|c|c|}
\hline
\multirow{2}{*}{\textbf{Batch Size}} & \multirow{2}{*}{\textbf{Points}} &
\multirow{2}{*}{\textbf{Model}} & \multicolumn{2}{|c|}{\textbf{$50 \times 50$}} & 
\multicolumn{2}{|c|}{\textbf{$100 \times 100$}} \\
\cline{4-7}
{} & {} & {} & {MSE} & {Rel. Error} & {MSE} & {Rel. Error} \\
\hline
\multirow{4}{*}{$500$} & \multirow{2}{*}{[C]} & {\textit{BCXN-CAN}} & {\scriptsize$2.63 \times 10^{-4}$} & {\scriptsize$7.75 \times 10^{-2}$} & {\scriptsize$3.06 \times 10^{-3}$} & {\scriptsize$2.63 \times 10^{-1}$} \\
\cline{3-7}
{} & {} & {\textit{CAN \& AD}} & {\tiny$>1 \times 10^{-2}$} & {\tiny$>5 \times 10^{-1}$} & {\tiny$>1 \times 10^{-2}$} & {\tiny$>5 \times 10^{-1}$} \\
\cline{2-7}
{} & \multirow{2}{*}{[N]} & {\textit{CAN}} & {\tiny$>1 \times 10^{-2}$} & {\tiny$>5 \times 10^{-1}$} & {\tiny$>1 \times 10^{-2}$} & {\tiny$>5 \times 10^{-1}$} \\
\cline{3-7}
{} & {} & {\textit{AD}} & {\tiny$>1 \times 10^{-2}$} & {\tiny$>5 \times 10^{-1}$} & {\tiny$>1 \times 10^{-2}$} & {\tiny$>5 \times 10^{-1}$} \\
\hline
\multirow{4}{*}{$1000$} & \multirow{2}{*}{[C]} & {\textit{BCXN-CAN}} & {\scriptsize$2.23 \times 10^{-4}$} & {\scriptsize$7.13 \times 10^{-2}$} & {\scriptsize$3.97 \times 10^{-4}$} & {\scriptsize$9.47 \times 10^{-2}$} \\
\cline{3-7}
{} & {} & {\textit{CAN \& AD}} & {\tiny$>1 \times 10^{-2}$} & {\tiny$>5 \times 10^{-1}$} & {\tiny$>1 \times 10^{-2}$} & {\tiny$>5 \times 10^{-1}$} \\
\cline{2-7}
{} & \multirow{2}{*}{[N]} & {\textit{CAN}} & {\tiny$>1 \times 10^{-2}$} & {\tiny$>5 \times 10^{-1}$} & {\tiny$>1 \times 10^{-2}$} & {\tiny$>5 \times 10^{-1}$} \\
\cline{3-7}
{} & {} & {\textit{AD}} & {\tiny$>1 \times 10^{-2}$} & {\tiny$>5 \times 10^{-1}$} & {\tiny$>1 \times 10^{-2}$} & {\tiny$>5 \times 10^{-1}$} \\
\hline
\multirow{4}{*}{$2000$} & \multirow{2}{*}{[C]} & {\textit{BCXN-CAN}} & {\scriptsize$2.36 \times 10^{-4}$} & {\scriptsize$7.34 \times 10^{-2}$} & {\footnotesize$7.14 \times 10^{-5}$} & {\footnotesize$4.02 \times 10^{-2}$} \\
\cline{3-7}
{} & {} & {\textit{CAN \& AD}} & {\tiny$>1 \times 10^{-2}$} & {\tiny$>5 \times 10^{-1}$} & {\tiny$>1 \times 10^{-2}$} & {\tiny$>5 \times 10^{-1}$} \\
\cline{2-7}
{} & \multirow{2}{*}{[N]} & {\textit{CAN}} & {\scriptsize$7.41 \times 10^{-4}$} & {\scriptsize$1.32 \times 10^{-1}$} & {\tiny$>1 \times 10^{-2}$} & {\tiny$>5 \times 10^{-1}$} \\
\cline{3-7}
{} & {} & {\textit{AD}} & {\tiny$>1 \times 10^{-2}$} & {\tiny$>5 \times 10^{-1}$} & {\tiny$>1 \times 10^{-2}$} & {\tiny$>5 \times 10^{-1}$} \\
\hline
\multirow{4}{*}{$4000$} & \multirow{2}{*}{[C]} & {\textit{BCXN-CAN}} & {\cellcolor{lightgray}} & {\cellcolor{lightgray}} & {\footnotesize$2.38 \times 10^{-5}$} & {\footnotesize$2.31 \times 10^{-2}$} \\
\cline{3-7}
{} & {} & {\textit{CAN \& AD}} & {\cellcolor{lightgray}} & {\cellcolor{lightgray}} & {\tiny$>1 \times 10^{-2}$} & {\tiny$>5 \times 10^{-1}$} \\
\cline{2-7}
{} & \multirow{2}{*}{[N]} & {\textit{CAN}} & {\cellcolor{lightgray}} & {\cellcolor{lightgray}} & {\scriptsize$2.94 \times 10^{-4}$} & {\scriptsize$8.24 \times 10^{-2}$} \\
\cline{3-7}
{} & {} & {\textit{AD}} & {\cellcolor{lightgray}} & {\cellcolor{lightgray}} & {\tiny$>1 \times 10^{-2}$} & {\tiny$>5 \times 10^{-1}$} \\
\hline
\end{tabular}
\label{tab: ldc-sampling-effect}
\end{adjustbox}
\end{center}
\end{table}

From this experiment, we notice that the Node-centered \textit{CAN-loss} PINN model has improved performance relative to the Cell-centered version, for a larger batch size and training samples. This further emphasizes our hypothesis that accurate modelling of the boundary conditions is an important factor in obtaining good performance. As this is a square domain, the use of a sampling point distribution that has perfect connectivity with the boundary of the domain naturally leads to improved performance. 

In addition, the inclusion of our proposed \textit{BCXN-loss} on a Cell-centered LSA-PINN can improve model performance to achieve a lower error than the properly aligned baseline PINN models, especially for sparse training samples setting. This suggests that the use of LSA-PINN can afford flexibility in the choice of sampling strategies, whereas conventional sampling strategies may require sampling that maintains connectivity with the boundaries.

\begin{figure}
\centering
\subfigure{\includegraphics[width=0.5\textwidth,keepaspectratio]{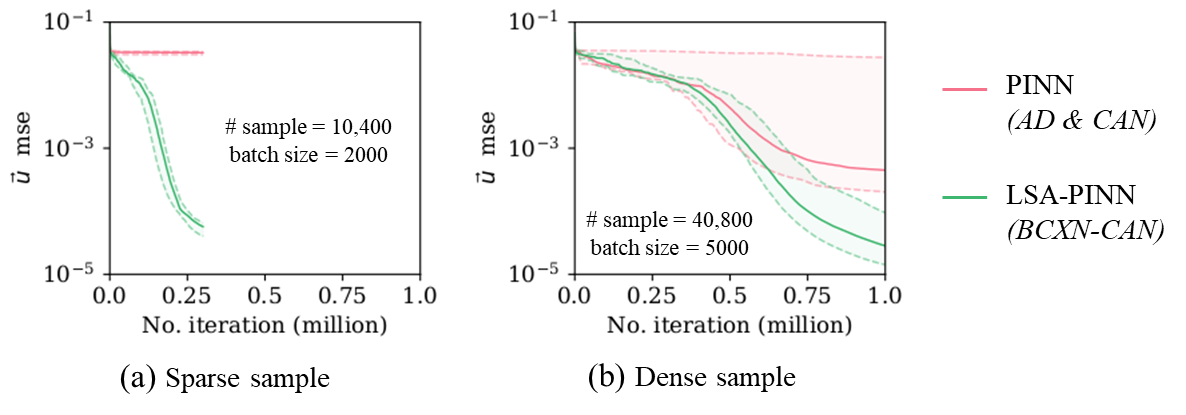}}
\vspace*{-7mm}
\caption{PINNs learning the solution of lid-driven cavity flow under (a) sparse and (b) dense training scenarios respectively. The convergence trends of $\vec{u}$ MSE are plotted. Bold lines indicate the median convergence path, and the shaded areas indicate the inter-percentile range from $10^{th}-90^{th}$ percentiles.}
\label{fig:convergence-ldc}
\vspace*{-5mm}
\end{figure}

\textbf{Case study with different sampling density.} From the above experiment, we notice that a 4x increase in sampling density (from $50 \times 50$ to $100 \times 100$) can improve model performance. Hence, we further study the lid-driven cavity flow problem to understand the trade-off between convergence in accuracy and sampling density. We train on a denser (i.e., $200 \times 200$) set of samples until the accuracy of the \textit{CAN} and \textit{AD} models start to improve. Note that denser sampling requires larger training iteration and batch size, and the tuning of these training hyper-parameters can be very time consuming. Although all PINN models can eventually achieve improved accuracy with denser sampling, Fig.~\ref{fig:convergence-ldc} indicates that it is much harder and slower to learn from a dense set of samples, thereby demonstrating the benefits of LSA-PINN. Encouragingly, LSA-PINN learns more accurate solutions ($>$1 order of magnitude lower $\vec{u}$ MSE) than the baseline \textit{CAN-} and \textit{AD-loss} PINNs with less ($<$50\%) training iterations, while using fewer ($<$50\%) training samples for the lid-driven cavity flow problem. 

\subsection{CNN-architecture LSA-PINNs} \label{cnn-fast}
The \textit{BCXN-loss} method does not pose any requirement on the differentiable property of the networks, hence it can be freely implemented in any neural network architecture, including CNN. Hence, we compare the LSA-PINN \textit{(BCXN-ND)} to baseline PINN \textit{(ND)} method previously published for the CNN architecture and summarize their average MSE and relative L2 error across multiple independent runs in Table \ref{tab: cnn-error-all}. The U-Net architecture and training settings used are summarized in Table~\ref{tab02}. All PINN models for this section are trained on a single discretized 2D input which consists of 3 channels, namely the x-, y-coordinate, and an indicator to differentiate fluid and non-fluid domain. They were trained for 200,000 iterations using Adam algorithm on a relatively sparse set of training samples, i.e., the semi-circle and square lid-driven cavity flow problems use 3,930 and 10,000 interior points which belong to the fluid domain respectively. 

\begin{table}[htbp]
\caption{Model Performance Comparisons for CNN-architecture PINNs}
\begin{adjustbox}{width=.48\textwidth}
\begin{tabular}{|c|c|c|c|}
\hline
\textbf{Problem} & \textbf{Model} & \textbf{MSE} & \textbf{Rel. Error} \\
\hline
\multirow{2}{*}{\parbox{3cm}{\centering Semi-circle lid-driven cavity flow, $Re = 1000$}} & {\textit{BCXN-ND}} & {$1.14 \times 10^{-4}$} & {$4.92 \times 10^{-2}$} \\
\cline{2-4}
{} & {\textit{ND}} & {$3.96 \times 10^{-2}$} & {$8.55 \times 10^{-1}$} \\
\hline
\multirow{2}{*}{\parbox{3cm}{\centering Lid-driven cavity flow, $Re = 1000$}} & {\textit{BCXN-ND}} & {$2.52 \times 10^{-5}$} & {$4.46 \times 10^{-2}$} \\
\cline{2-4}
{} & {\textit{ND}} & {$2.38 \times 10^{-2}$} & {$1.00 \times 10^{0}$} \\
\hline
\multirow{2}{*}{\parbox{4cm}{\centering Transient semi-circle lid-driven cavity flow, $Re = 500$}} & {\textit{BCXN-ND}} & {$3.95 \times 10^{-4}$} & {$1.29 \times 10^{-1}$} \\
\cline{2-4}
{} & {\textit{ND}} & {$9.66 \times 10^{-3}$} & {$6.81 \times 10^{-1}$} \\
\hline
\end{tabular}
\label{tab: cnn-error-all}
\end{adjustbox}
\end{table}

Like the MLP experiment, LSA-PINN for CNNs show a noticeable improvement in the solution accuracy (2-3 orders of magnitude lower $\vec{u}$ MSE) over the baseline \textit{ND-loss} PINN. The results indicate that LSA-PINN can effectively learn the solution of fluid dynamic problems with U-Net architecture and that the \textit{BCXN-loss} can benefit CNN-architecture PINNs by remedying training issues stemming from a discretized input being unable to exactly resolve complex geometries.

\textbf{Transient 2D semi-circle lid-driven cavity flow, $Re=500$.} In addition to the steady-state semi-circle lid-driven cavity flow, we further evaluate the improvement obtained from application of CNN-architecture LSA-PINN to the modelling of a transient version of this problem. In this instance, the problem's initial condition corresponds to that of a fluid domain at rest (at t = 0). The PINN model is then used to model the evolution of the flow within a semi-circle lid-driven cavity from $t = 0$ to $t = 2$, $\Delta{t} = 0.1$. Model settings are provided in Table~\ref{tab02}. The transient PINN models have additional $t$ channel in their input.

\begin{figure}[htbp]
\begin{center}
\centerline{\includegraphics[width=0.5\textwidth,keepaspectratio]{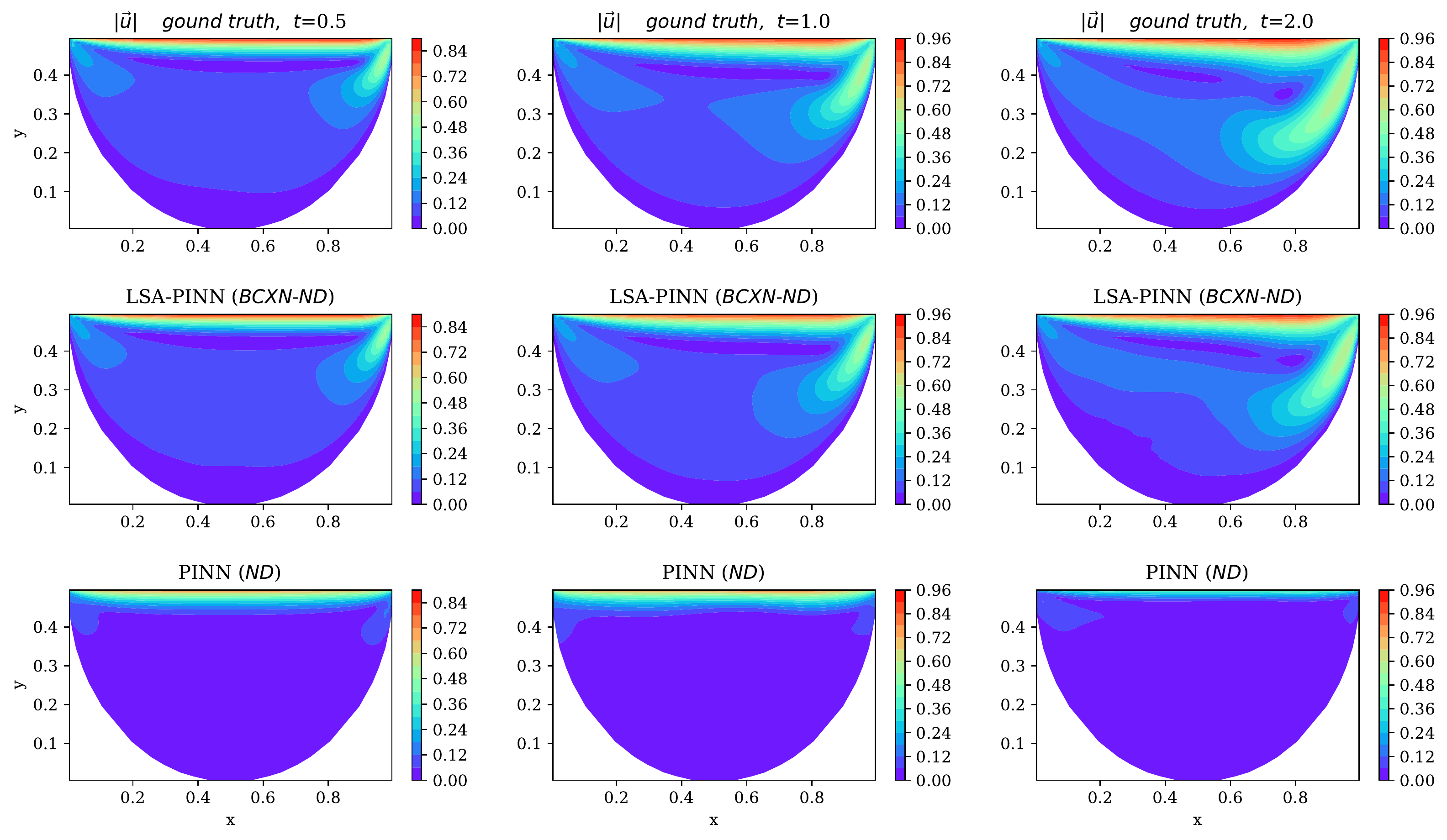}}
\caption{Velocity contours of the transient simulation at $t = 0.5$, $t = 1.0$ and $t=2.0$ for the ground truth and PINN solutions.}
\label{fig:transient-contour}
\end{center}
\vspace*{-6mm}
\end{figure}

The performance of LSA-PINN \textit{(BCXN-ND)} and baseline PINN \textit{(ND)} are compared in Table \ref{tab: cnn-error-all}. The average errors for all time points between $t = 0$ and $t = 2.0$ yield MSEs of 3.95e-4 and 9.66e-3 with and without the inclusion of \textit{BCXN-loss} respectively. Hence, the inclusion of \textit{BCXN-loss} in this transient problem improves MSE by $>1$ order of magnitude. Velocity contour plots of the \textit{BCXN-ND} and \textit{ND} models are provided in Fig. \ref{fig:transient-contour} to illustrate the differences in convergence.

\textbf{Performance with increasing problem complexity.} To better understand the improvement in accuracy obtainable by the \textit{BCXN-loss}, we also investigate the utility of LSA-PINN for modelling more complex problems more efficiently. Hence, we train both LSA-PINN and baseline PINN for the semi-circle lid-driven cavity flow across a range of $Re$. As $Re$ increases, the complexity of the developed flow typically requires more sophisticated algorithms and training samples for better capture of the non-linear effects.

The average MSE and relative L2 error obtained across multiple runs for various $Re$ are presented in Table \ref{tab: semi-ldc-table}. We clearly see that the performance difference is increasingly pronounced for more complicated (higher $Re$) problems, thus demonstrating that strategies like LSA-PINN methods are increasingly important for more complex problems.

\begin{table}[htbp]
\caption{Model Performance for the 2D Semi-circular Lid-Driven Cavity Problem}
\begin{adjustbox}{width=.48\textwidth}
\begin{tabular}{|c|c|c|c|c|c|c|}
\hline
\multirow{2}{*}{\textbf{Metric}} & \multirow{2}{*}{\textbf{Model}} & \multicolumn{5}{|c|}{\textbf{Reynolds Number (\textit{Re})}} \\
\cline{3-7}
{} & {} & {$50$} & {$100$} & {$200$} & {$400$} & {$1000$} \\
\hline
\multirow{2}{*}{MSE} & {\textit{BCXN-ND}} & {$2.51 \times 10^{-5}$} & {$2.76 \times 10^{-5}$} & {$3.56 \times 10^{-5}$} & {$4.58 \times 10^{-5}$} & {$1.14 \times 10^{-4}$}\\
\cline{2-7}
{} & {\textit{ND}} & {$2.98 \times 10^{-4}$} & {$2.79 \times 10^{-4}$} & {$3.74 \times 10^{-4}$} & {$2.33 \times 10^{-3}$} & {$3.96 \times 10^{-2}$} \\
\hline
\multirow{2}{*}{Rel. Error} & {\textit{BCXN-ND}} & {$0.0258$} & {$0.0269$} & {$0.0305$} & {$0.0335$} & {$0.0492$} \\
\cline{2-7}
{} & {\textit{ND}} & {$0.0823$} & {$0.0732$} & {$0.0780$} & {$0.211$} & {$0.855$} \\
\hline
\end{tabular}
\label{tab: semi-ldc-table}
\end{adjustbox}
\end{table}

\begin{figure}[H]
\centering
\subfigure{\includegraphics[width=\linewidth,keepaspectratio]{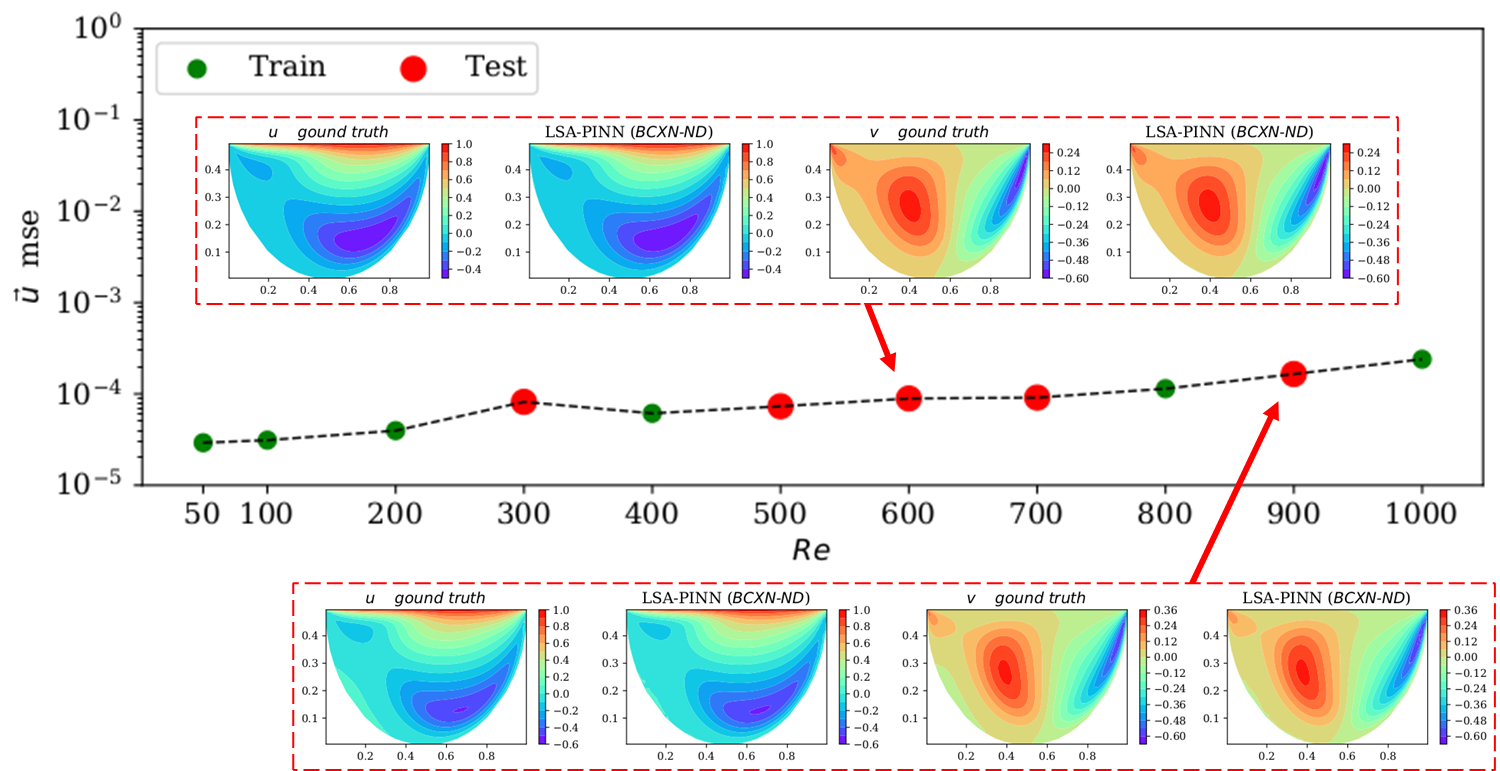}}
\vspace*{-6mm}
\caption{LSA-PINN learning the solutions of multiple 2D semi-circle lid-driven cavity flow scenarios concurrently, using CNN architecture. The $\vec{u}$ MSE between ground truth and LSA-PINN solution for train and test $Re$ are plotted. Inset are $u$- and $v$-velocity contours of the ground truth and LSA-PINN solutions for $Re=600$ and $Re=900$.}
\label{fig:meta-ldc}
\end{figure}

\textbf{Meta-PINN model.} The CNN-architecture PINN models represent the entire spatial domain as a single input. We can train the model on multiple samples, each representing a different scenario such as changes in geometry or physical behaviour. Such a meta-PINN model can be used to predict new scenarios. Hence, we further study the ability of LSA-PINN to learn the solutions to multiple flow scenarios concurrently (spanning $Re=50-1000$) on semi-circle lid-driven cavity geometry. The model architecture and training setting is given in Appendix \ref{table-settings} Table~\ref{tab02}. The LSA-PINN \textit{(BCXN-ND)} model is used to predict $Re=300, 500, 600, 700, 900$ scenarios which are not used during training. The results are summarized in Fig.~\ref{fig:meta-ldc}. The present LSA-PINN method successfully learns a good solution with low $\vec{u}$ MSE (i.e., $<$1e-4) for most of the $Re$ scenarios, and achieves good prediction accuracy.

\subsection{Impact of \textit{BCXN}-loss on training time} \label{training-time}
We further compare the time taken to train a model with and without the \textit{BCXN-loss} on the MLP- and CNN-type models as applied to the semi-circle lid-driven cavity problem at $Re = 1000$. The average MSE and relative L2 error obtained are averaged across 10 independent runs, and the minimum and maximum wall-clock time are recorded. All experiments are run with standard Tensorflow and Keras implementation on a 20-core workstation with an Intel Xeon Gold 6248 processor and an Nvidia RTX 2080 Ti GPU.

Quantitative comparisons for the MLP-architecture and CNN-architecture models are presented in Table \ref{tab: mlp-error-time}. For fair comparison within architecture types, all MLP models were run for an identical number of training iterations (300,000) with the same batch size (1,000) and model size while all CNN models were similarly run for an identical number of training iterations (200,000) and with the same model size.

\begin{table}[htbp]
\caption{Model Performance for different PINN architectures}
\begin{center}
\begin{tabular}{|c|c|c|c|c|}
\hline
\textbf{Model} & \textbf{Method} & \textbf{MSE} & \textbf{Rel. Error} & \textbf{Time (min)} \\
\hline
{MLP} & {\textit{BCXN-CAN}} & {$6.21 \times 10^{-5}$} & {$3.55 \times 10^{-2}$} & {$55-57$} \\
\hline
{MLP} & {\textit{BCXN-ND}} & {$1.61 \times 10^{-4}$} & {$5.95 \times 10^{-2}$} & {$23-24$} \\
\hline
{MLP} & {\textit{CAN}} & {$2.82 \times 10^{-2}$} & {$7.62 \times 10^{-1}$} & {$40-46$} \\
\hline
{MLP} & {\textit{ND}} & {$3.91 \times 10^{-2}$} & {$8.56 \times 10^{-1}$} & {$20-22$} \\
\hline
{MLP} & {\textit{AD (SIREN)}} & {$1.74 \times 10^{-2}$} & {$5.95 \times 10^{-1}$} & {$58-61$} \\
\hline
{MLP} & {\textit{AD} (Vanilla)} & {$5.46 \times 10^{-2}$} & {$9.83 \times 10^{-1}$} & {$50-55$} \\
\hline
{CNN} & {\textit{BCXN}} & {$1.14 \times 10^{-4}$} & {$4.92 \times 10^{-2}$} & {$126-136$} \\
\hline
{CNN} & {\textit{ND}} & {$3.96 \times 10^{-2}$} & {$8.55 \times 10^{-1}$} & {$121-137$} \\
\hline
\end{tabular}
\label{tab: mlp-error-time}
\end{center}
\end{table}

From this study, we note that the inclusion of \textit{BCXN-loss} across all the cases only slightly increases the amount of time required for model training. This increase was on the order of less than 5 \% for the CNN-type architecture and the ND-MLP-type model. Critically, while the time taken may have increased slightly, the MSE typically decreases more than 1 order of magnitude. In many instances, this trade-off between model training time and performance with the inclusion of a \textit{BCXN-loss} is probably preferable.

\section{Conclusion}
We present a LSA-PINN method which can more efficiently learn the solution to PDEs for complex geometry with sparse training samples. This is accomplished by enforcing a linear constraint during the training implicitly. While linear constraints are applied in this work due to truncation of the Taylor series expansion in the derivation, it is worth studying if different, higher order approximations can achieve better performance in future work. Overall, our comprehensive experimental studies demonstrate practical advantages of LSA-PINN on diverse test problems in fluid dynamics, improving the accuracy of solutions by orders of magnitude even while requiring much less training iterations and samples. 


\bibliographystyle{IEEEtran}
\bibliography{IEEEabrv,ref_bib}{}

\appendix
\section{Appendix}

\subsection{MLP and CNN experimental settings} \label{table-settings}

\begin{table*}[htbp]
\centering
\caption{PINN (MLP) architecture and training settings used for experiments in Section \ref{mlp-pinn}}
\begin{adjustbox}{width=.55\textwidth}
\begin{tabular}{LLLL}\toprule
    \textbf{Problem type} & \textbf{Forward}  & \textbf{Forward} & \textbf{Forward} \\ \hline
    Test case                & 2D semi-circle lid-driven cavity flow, $Re=1000$ & 2D lid-driven cavity flow, $Re=1000$ & 2D wavy channel flow, $Re=100$ \\\\
    \textit{Governing physics eqns.} & 2D \textit{incompressible NS eqns.} & 2D \textit{incompressible NS eqns.} & 2D \textit{incompressible NS eqns.} \\ \hline
    MLP architecture \textit{for LSA-PINN, ND-PINN $\&$ AD-PINN} & $(x,y)-64–30–30–30–$ $[30–30–30–(\hat{u})$, $30–30–30–(\hat{v})$, $30–30–30–(\hat{p})]$ & $(x,y)-128–30–30–30–$ $[30–30–30–(\hat{u})$, $30–30–30–(\hat{v})$, $30–30–30–(\hat{p})]$ & $(x,y)-128–40–40–40–$ $[40–40–40–(\hat{u})$, $40–40–40–(\hat{v})$, $40–40–40–(\hat{p})]$ \\
    ND scheme \textit{for LSA-PINN $\&$ ND-PINN} & CAN \& finite difference & CAN & CAN \\\\
    Training sample \textit{(equidistantly spaced samples @ fluid domain $\&$ boundary)} & $3,930+300$ & $10,000 + 400$ & $14,400 + 800$\\\\
    Batch size & $1,000$ & $2,000$ & $1,000$ \\\\
    Training iterations & $300,000$ & $300,000$& $300,000$\\
\bottomrule
\end{tabular}
\end{adjustbox}
\begin{enumerate}[(I)]

\item 
For the MLP architecture, the numbers in between input and output represent the number of nodes in each hidden layer. For example, $(x)–64–20–20–20–(\hat{u})$ indicates a single input $x$, followed by 4 hidden layers with 64, 20, 20 and 20 nodes in each layer, and a single output $\hat{u}$.
\item 
We incorporate the sinusoidal mapping ~\cite{wong2022learning} into the first hidden layer of PINN and initialize its weights by sampling from a normal distribution $N(0,\sigma^2), \sigma=1$. The subsequent hidden layers use “sine” activation, except a “linear” activation function is used in the final (output) layer, and their weights are initialized by He uniform distribution.
\item
Batch size: number of random sample used for 1 evaluation of $L_{PINN} (=\lambda_{PDE}~ L_{PDE}+\lambda_{BC}~L_{BC})$ and $L_{PINN(df-BCXN)}  (=\lambda_{PDE}~L_{PDE(df-BCXN)} )$. We used a default $\lambda_{PDE}=1$ and $\lambda_{BC}=1$ unless otherwise mentioned.
\item
A training iteration: 1 evaluation of $L_{PINN}$ or  $L_{PINN(df-BCXN)}$ for backpropagating the weight gradients. We use an initial learning of 1e-3 and reduce it on plateauing, until a min. learning rate of 5e-6 is reached.
\end{enumerate}
\label{tab01}
\end{table*}

\begin{table*}[t]
\centering
\caption{PINN (CNN) architecture and training settings used in the experiments in Section \ref{cnn-fast}}
\begin{adjustbox}{width=.85\textwidth}
\begin{tabular}{RRRRR}\toprule
    \textbf{Problem type} & \textbf{Forward}  & \textbf{Forward} & \textbf{Forward} & \textbf{Meta-modelling} \\ \hline
    Test case                & 2D semi-circle lid-driven cavity flow, $Re=1000$ & 2D lid-driven cavity flow, $Re=1000$ & 2D transient semi-circular lid-driven cavity flow, $Re=500$ & 2D semi-circle lid-driven cavity flow,$Re=1000$ \\\\
    \textit{Governing physics eqns.} & 2D \textit{incompressible NS eqns.} & 2D \textit{incompressible NS eqns.} & 2D \textit{transient incompressible NS eqns.} & 2D \textit{incompressible NS eqns.} \\ \hline
    CNN (U-Net) architecture \textit{for LSA-PINN $\&$ ND-PINN} & $(\textit{\textbf{X}}_{(56\times112\times3)})–$,
$8–8–\downarrow–16–16–\downarrow–$,
$32–32–\uparrow–64–64–\uparrow–$,
$32–32–\uparrow–16–16–\uparrow–[$,
$8–8–8–(\hat{\textit{\textbf{U}}}_{u,56\times112})$,
$8–8–8–(\hat{\textit{\textbf{U}}}_{v,56\times112})$,
$8–8–8–(\hat{\textit{\textbf{U}}}_{p,56\times112})$ & $(\textit{\textbf{X}}_{(104\times104\times3)})–$,
$8–8–\downarrow–16–16–\downarrow–$,
$32–32–\uparrow–64–64–\uparrow–$,
$32–32–\uparrow–16–16–\uparrow–[$,
$8–8–8–(\hat{\textit{\textbf{U}}}_{u,104\times104})$,
$8–8–8–(\hat{\textit{\textbf{U}}}_{v,104\times104})$,
$8–8–8–(\hat{\textit{\textbf{U}}}_{p,104\times104})$ &$(\textit{\textbf{X}}_{(56\times112\times4)})–$,
$8–8–\downarrow–16–16–\downarrow–$,
$32–32–\uparrow–64–64–\uparrow–$,
$32–32–\uparrow–16–16–\uparrow–[$,
$8–8–8–(\hat{\textit{\textbf{U}}}_{u,56\times112})$,
$8–8–8–(\hat{\textit{\textbf{U}}}_{v,56\times112})$,
$8–8–8–(\hat{\textit{\textbf{U}}}_{p,56\times112})$ & $(\textit{\textbf{X}}_{(56\times112\times4)})–$,
$8–8–\downarrow–16–16–\downarrow–$,
$32–32–\uparrow–64–64–\uparrow–$,
$32–32–\uparrow–16–16–\uparrow–[$,
$8–8–8–(\hat{\textit{\textbf{U}}}_{u,56\times112})$,
$8–8–8–(\hat{\textit{\textbf{U}}}_{v,56\times112})$,
$8–8–8–(\hat{\textit{\textbf{U}}}_{p,56\times112})$ \\\\
    Kernel size & $4\times4$ & $6\times6$ & $4\times4$ & $6\times6$ \\\\
    ND scheme \textit{for LSA-PINN $\&$ ND-PINN} & Finite difference & Finite difference & Finite difference & Finite difference \\\\
    Training sample \textit{~~~~~~~~/ batch size} & 1 / 1 & 1 / 1 & 21 / 2 & 6 / 1 \\\\
    Training iterations & $200,000$ & $200,000$& $100,000$ & $500,000$ \\
\bottomrule
\end{tabular}
\end{adjustbox}
\begin{enumerate}[(I)]

\item 
For the U-net architecture, the numbers in between input and output represent the number of filters in each hidden layer. Each hidden layer consists of convolution operation, batch normalization, and nonlinear “sine” activation. Also, $\downarrow$ represents a $2\times2$ max-pooling operation, and $\uparrow$ represents a $2\times2$ up-sampling operation. The network weights are initialized by He uniform distribution.
\item 
Batch size: number of sample used for 1 evaluation of $L_{PINN} (=\lambda_{PDE}~L_{PDE}+\lambda_{BC}~L_{BC})$ and $L_{PINN(df-BCXN)}  (=\lambda_{PDE}~L_{PDE(df-BCXN)} )$. We used a default $\lambda_{PDE}=1$ and $\lambda_{BC}=1$, unless otherwise mentioned.
\item
A training iteration: 1 evaluation of $L_{PINN}$ or $L_{PINN(df-BCXN)}$ for backpropagating the weight gradients. We use an initial learning of 1e-3 and reduce it on plateauing, until a min. learning rate of 5e-6 is reached.
\item The semi-circle and regular lid-driven cavity geometries consist of 3,930 and 10,000 interior points which belong to the fluid domain respectively.
\end{enumerate}
\label{tab02}
\end{table*}

\end{document}